\documentclass[accepted]{uai2021} 

\usepackage[american]{babel}

\usepackage{natbib} 
\usepackage{mathtools} 
\usepackage{booktabs} 
\usepackage{tikz} 

\usepackage{amsmath,amsthm,verbatim,amssymb,amsfonts,amscd, graphicx}
\usepackage{graphics}
\usepackage{centernot}

\theoremstyle{plain}
\newtheorem{theorem}{Theorem}
\newtheorem{corollary}{Corollary}

\newtheorem{proposition}{Proposition}

\theoremstyle{definition}

\title{On the Distributional Properties of Adaptive Gradients}

\author[1]{\href{mailto:Zhang Zhiyi <zhiyi.stat@gmail.com>?Subject=Your UAI 2021 paper}{Zhang Zhiyi}{}}
\author[2]{\href{mailto:Liu Ziyin <zliu@cat.phys.s.u-tokyo.ac.jp>?Subject=Your UAI 2021 paper}{Liu Ziyin}{}}

\affil[1]{%
    School of Statistics, Xi’an University of Finance and Economics
}
\affil[2]{%
    Department of Physics, University of Tokyo
}

\begin{document}
\maketitle

\begin{abstract}
    Adaptive gradient methods have achieved remarkable success in training deep neural networks on a wide variety of tasks. However, not much is known about the mathematical and statistical properties of this family of methods. This work aims at providing a series of theoretical analyses of its statistical properties justified by experiments. In particular, we show that when the underlying gradient obeys a normal distribution, the variance of the magnitude of the \textit{update} is an increasing and bounded function of time and does not diverge. This work suggests that the divergence of variance is not the cause of the need for warm up of the Adam optimizer, contrary to what is believed in the current literature.
    
\end{abstract}

\vspace{-2mm}
\section{Introduction}
\vspace{-1mm}

In the last ten years, the optimization of deep neural networks has become an important research topic \citep{Zhang_optimizationProperitesSGD, Im_empiricalAnalysisoftheOptimization, le2011optimization, choi2020empirical, sun2019optimization, barakat2018convergence, barakat2020convergence}. Designing larger and larger neural networks puts an increasing demand for developing an efficient neural network training algorithm. Traditionally, stochastic gradient descent is deployed to train neural networks. In the last ten years, the adaptive gradient family, including but not limited to RMSProp \citep{Tieleman2012_rmsprop}, Adam \citep{journals/corr/KingmaB14_adam}, AdaGrad \citep{Duchi:2011:ASM:1953048.2021068}, has emerged as the major tool for training deep neural networks. Many variants in the adaptive gradient family have been proposed, but none of these methods has shown dominating advantage or popularity over the other \citep{Reddi2018convergence, liu2019variance, loshchilov2017fixing, Luo2019AdaBound}. Two notable works that advanced mathematical understanding of the adaptive gradients include the work by \citep{Reddi2018convergence}, which shows that, when hyperparameter does not match the setting of the problem, the adaptive gradient method might not converge at all, and the work by \citep{liu2019variance}, which proposes a new algorithm based on the argument that, at initialization steps, the adaptive gradient method has a divergent variance.

In this work, we take the first step for studying a rather fundamental problem in the study of adaptive gradients; we propose to study the distributional properties of the update in the adaptive gradient method. The most closely related previous work is \citep{liu2019variance}. The difference is that this work goes much deeper into the detail in the theoretical analysis and contradicts the results in \citep{liu2019variance}. The main contributions of this work are the following: (1) We prove that the variance of the adaptive gradient method is always finite (Proposition 1), which contradicts the result in \cite{liu2019variance}; this proof does not make any assumption regarding the distribution of the gradient. (2) Under the assumption that the gradient is time-independent isotropic Gaussian and that the preconditioner $n_t$ is obtained by merely averaging (same as in previous work \cite{liu2019variance}), we derive the exact distribution of the update for every time-step $t$. While the derivation is simple, the exact formula is not known previously (\textit{Section 3-4}). (3) The predicted distribution is shown to agree well with experiments, even on modern architectures, including the transformers, with state-of-the-art performance and a single training trajectory level (\textit{Section 5 and Supplementary}); (4) We experimentally study and discuss when and why experiments could deviate from our theoretical prediction (\textit{Section 6}).
\begin{figure}
    \centering
    \includegraphics[width=0.8\linewidth]{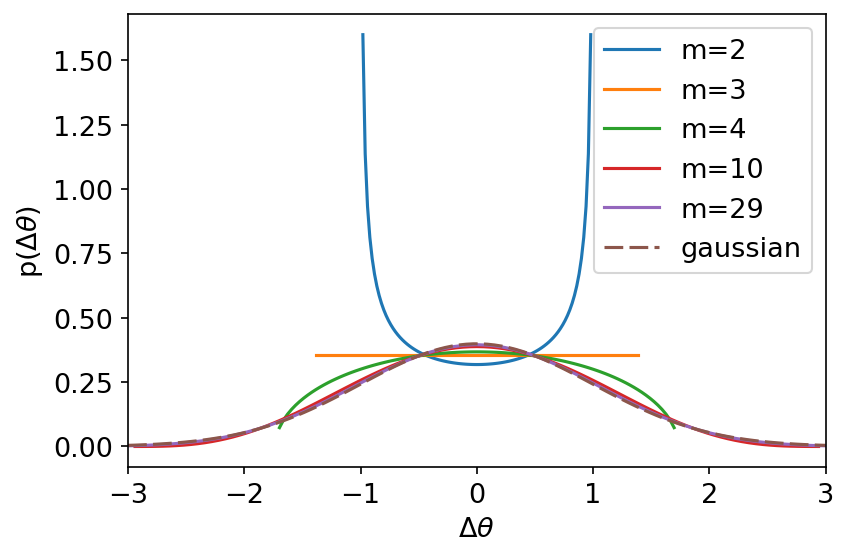}
    \vspace{-1em}
    \caption{Predicted distribution of the update $\Delta \theta$ of the Adam algorithm at a given time step $t=m$. It has the following interesting property: (1) the distribution is bounded for any finite $m$; (2) the variance equals $1$ for all $m$; (3) it transitions from a bi-modal to unimodal distribution as $m$ increases; (4) as $m\to \infty$, the distribution converges to a Gaussian distribution.}
    \vspace{-1em}
    \label{fig: theo pdf}
\end{figure}

\vspace{-2mm}
\section{Related works}
\vspace{-1mm}
\textbf{Adaptive learning rate methods.}
The adaptive gradient methods have emerged as the most popular tool for training deep neural networks over the last few years, and they have been of great use both industrially and academically. The adaptive gradient family makes an update by dividing the gradient by the running root mean square (RMS) of the gradients \citep{Duchi:2011:ASM:1953048.2021068, Tieleman2012_rmsprop, journals/corr/KingmaB14_adam}, which speeds up training by effectively rescaling the update to the order of $O(1)$ throughout the training trajectory. The most popular method among this family is the Adam algorithm \citep{journals/corr/KingmaB14_adam}, which computes the momentum and the preconditioner as exponential averages with decay hyperparameter $\mu,\ \beta$ (also referred to as $\beta_1,\ \beta_2$ in literature), where bias correction terms $c_n(t), c_m(t)$ correct the bias in the initialization. The Adam algorithm can be written as:
\begin{align}\label{eq: laprop begin}
    g_{t} &= \nabla_\theta \ell(\theta_{t-1}), \\
    n_t &= \beta n_{t-1} + (1-\beta) g_{t}^2, \\
    m_t &= \mu m_{t-1} + (1-\mu) \frac{g_{t}}{\sqrt{n_t/c_n} + \epsilon},  \\
    \theta_t &= \theta_{t-1} - \lambda \frac{ m_t }{c_m}; \label{eq: laprop update}
\end{align}
$g_t$ is the gradient at time step $t$; Adam sets $c_n(t)={{1 - \beta^t}}$ and $c_m(t)={{1 - \mu^t}}$; in the literature, $1/\sqrt{n_t/c_n}$ is called the \textit{preconditioner}; and $\epsilon$ is a very small numerical bias to prevent divergence. This work focuses on the study of the distribution of the \textit{update}, defined as $\Delta \theta :=(\theta_t - \theta_{t-1})/\lambda$.
Several variants of Adam also exist \citep{Reddi2018convergence, liu2019variance, loshchilov2017fixing, Luo2019AdaBound}. However, it remains inconclusive as to which method is the best and the various ``fixes" of Adam do not show consistent better performance. Therefore, we focus on studying the original Adam algorithm, and we believe that qualitatively similar results will carry over to other algorithms as well. Theoretically, we also work in the limit when $\mu=0$ to avoid notational overload, but we note that its effect can be incorporated in a rather straightforward way.

\textbf{The need for warmup in recent adaptive gradient method.} While simpler models can be trained with the adaptive gradient methods with ease without paying particular care to the learning rate, larger and modern models tend to need special tuning of the learning rate $\lambda$. One notable example is the transformer architecture, which is the current state-of-the-art model for language tasks, and, what is more, it is empirically found that transformer can only be trained with a scheme that makes $\lambda=\lambda_t$ an explicit function of time \citep{vaswani2017attention, young2018recent, devlin2018bert, lan2019albert}. In particular, one often increases $\lambda$ from a minimal value to a maximum value with linear or other power-law monotonic functions through the beginning time steps \citep{popel2018training}. This learning rate scheduling technique is called \textit{warmup}. However, the cause of the need for warmup is not yet understood; it is imaginable that a correct understanding of the adaptive gradient's problem will advance our understanding of deep neural network optimization and benefit the industry. In \citep{liu2019variance}, it is argued that the necessity of warmup may be due to the divergence of Adam's variance at initialization. However, the result of this work suggests that it is not the case and that the understanding of the reason for the need for warmup remains open.

\vspace{-1mm}
\vspace{-1mm}
\section{Notation and Preliminaries}
\vspace{-1mm}
While our theory focuses on deriving distributions of the functions of a simple Gaussian variable $X$, it is helpful to keep in mind that, for application, $X$ will be linked to the gradient, and its estimated second momentum $U$ will be linked to the preconditioner that is commonly in use in the adaptive gradient methods, and the number of samples $m$ will be linked to the time step $t$ of an optimization trajectory. 

Let $X_i$ be i.i.d. random variables (RV) drawn from a Gaussian distribution $\mathcal{N}(\mu, \sigma)$. 
We may want to estimate its mean by taking average of $n$ many samples $\Bar{X} = \frac{1}{n} \sum_i^n X_i$. $\Bar{X}$ is also Gaussian
with variance scaled by $1/n$. We also want to estimate the variance of $X$, through $m$ many samples. When $\mu=0$, the maximum likelihood estimator (MLE) is $S = \frac{1}{m} Q :=\frac{1}{m} \sum_i^m X_i^2$. The RV $Q / \sigma^2$ obeys a $\chi^2$ distribution with degree of freedom $m$, whose density is
\begin{equation}
    f_m(\chi^2) = \frac{1}{2^{m/2}\Gamma(m/2)} (\chi^2)^{m/2 - 1} e^{-\chi^2/2}
\end{equation}
Here $\Gamma(x)$ is the gamma function.
The properties of the $\chi^2$ distribution is well-known, including (1) sum of $\chi^2$ RVs is again a $\chi^2$ RV with its degree of freedom added; (2) when $m=2$, we obtain the exponential distribution; (3) when $m\to \infty$, the density converges to a Gaussian with mean $m$. The empirical variance $S$ follows the \textit{reduced} $\chi^2$ distribution, whose density we can obtain by performing a change of variable $\chi^2 \to \chi^2/m$:
\begin{equation}
    f_m(s) = mf_m(\chi^2) =\frac{m}{2^{m/2}\Gamma(m/2)} (ms)^{m/2 - 1} e^{-ms/2}
\end{equation}
when $m$ is very large, this converges to a Gaussian with mean $1$ and variance $2/m$. Also, one might use the unbiased estimator $U := \frac{1}{n-1}Q$ instead of $S$ in many cases.

Now consider the RV $X$: we can transform it into a standard Gaussian by dividing by its standard deviation, $X \to X/\sigma$. However, when the true variance is not known, we have to divide by the estimated variance $\sqrt{U}$, and it can be fruitful to find the distribution for RV $T := X/\sqrt{U}$. Assuming that $U$ and $X$ are independent, the well-known result is that $t$ obeys the well-known student-$t$ distribution with the degree of freedom $m$.
When $m=1$, we obtain a Cauchy distribution, whose second moment diverges. The variance for this distribution is $\frac{m}{m-2}$. This means that we would like to estimate the denominator with $>2$ samples to avoid variance explosion. As $m \to \infty$, we see that the above distribution converges to a standard Gaussian, as expected. To be more general, we define
\[T := \frac{\Bar{X} -\mu}{\sqrt{U/m}}\]
and $T$ obeys a $t$-distribution with degree of freedom $m-1$. 

One might also consider a more general random variable. Let $X^2,\ Y^2$ be $\chi^2$ distributions with degree of freedom $m_1, m_2$ respectively, then we can define $Z:= \frac{X^2/m_1}{Y^2/m_2}$, and this follows the $F$-distribution.
We are interested in estimating the variable $\frac{(\Bar{X} - \mu)^2}{U^2/m}$. This obeys the $F_{1, m-1}$ distribution
\begin{equation}
    f_{1, m-1}(z) = \frac{(m-1)^{m-1/2} \Gamma(\frac{m}{2})}{\Gamma (\frac{1}{2}) \Gamma(\frac{m-1}{2})} \frac{z^{-1/2}}{(m-1 + z)^{m/2} }.
\end{equation}
Notice that if we consider the variable $\sqrt{z}$, then we recover the student's $t$-distribution. The $F_{1, m-1}$ distribution has mean $\frac{m-1}{m-3}$, and variance $\frac{2(m-1)^2(m-2)}{(m-2)^2(m-4)}$, which is a decreasing function of time (and is divergent when $M\leq 4$\footnote{Similar problem of the RAdam algorithm proposed in \citep{liu2019variance} has been noticed in \citep{ma2019adequacy}, where it is shown that, in most of the problems, the RAdam algorithm is equivalent to running $4$ steps of SGD and then switching to Adam. These two facts seem to be related.}). In the Theorem 1 of \citep{liu2019variance}, the gradient $X$ is assumed to be from a normal distribution, and the variable $Z$ is used to model the distribution of $\Delta \theta$, and it is shown that its variance decreases through time and diverges at the initial time steps. However, it is not hard to see that the update of Adam will not diverge when $0\leq\beta<1$ because the numerator $g_t$ and denominator $\sqrt{n_t/c_n}$ in the update are correlated, and this correlation suppresses the divergence, even if the bias factor $\epsilon=0$. One can show the following result.

\begin{proposition}\label{prop: bound}
When $\mu=0$, $0\leq\beta<1$, $\epsilon=0$, and when $g_t$ are i.i.d. gaussian variables, then $\Delta \theta$ is a sub-gaussian variable, whose higher moments exist and are bounded.
\end{proposition}
\textit{Proof sketch}. By the proposition 1 of \citep{ziyin2020laprop}, we have that when $\mu=0$, $0\leq\beta<1$, the update of Adam is bounded by a constant $c=c(\beta, t)$ that only dependent on $\beta$ and $t$. This means that $\Delta \theta$ is a bounded variable and, therefore, subgaussian. The boundedness of the higher moments follows from the established properties of a subgaussian variable. $\square$

This means that trying to separately understand the distribution of $g_t$ and $\sqrt{n_t/c_n}$ will lead to incorrect results, predicting divergence even if there is none. In the following discussion, we work out the actual distribution of $\Delta \theta$ using similar assumptions as in \citep{liu2019variance}. Contrary to the previous result, we show that the actual variance of $\Delta \theta$ is a constant in time while that of $|\Delta \theta|$ increasing function of time (instead of decreasing), and its distribution asymptotically converges to a gaussian as $m\to \infty$.

\vspace{-2mm}
\subsection{Effect of Exponential Averaging}
\vspace{-2mm}
In practice, we often resort to a form of exponential averaging in the deep learning optimization literature. Here the averaging of an RV $Z$ is defined as
$$\Bar{Z}_\beta = (1-\beta)\sum_{i=1}^m \beta^{m - i} Z_i$$
Often $0\leq \beta < 1$ to make the sum convergent, but the sum may be extrapolated to regions outside $[0, 1)$. By the additivity of Gaussian variables, the distribution is
\begin{equation}
    \Bar{Z}_\beta \sim \mathcal{N}\bigg(\mu (1 - \beta^{m-1}), \sigma^2  (1 -\beta )^2 \frac{1 - \beta^{2(m-1)}}{1 - \beta^2} \bigg) .
\end{equation}
We first note that, if $\beta \to 0$, then the distribution converges to $\mathcal{N} (\mu, \sigma^2)$. If $\beta \to 1$, then using L'Hopital's rule we find that its variance goes to $0$, i.e., converging to a delta distribution. To convert this distribution to an unbiased estimator of $\Bar{Z}$, we may divide $Z$ by $1-\beta^{m-1}$. Converting the distribution to
\begin{equation}
    \mathcal{N}\bigg(\mu, \sigma^2  \frac{(1 -\beta )^2}{(1 - \beta^{m-1})^2} \frac{1 - \beta^{2(m-1)}}{1 - \beta^2} \bigg) =  \mathcal{N}\bigg(\mu, \sigma^2  \frac{\sum_{i=0}^{m-1} \beta^{2i}}{(\sum_{i=0}^{m-1} \beta^{i})^2} \bigg)
\end{equation}
or, equivalently,
\begin{equation}
    \mathcal{N}\bigg(\mu, \sigma^2  \frac{1 -\beta}{1 + \beta} \frac{1 + \beta^{m-1}}{1 - \beta^{m-1}} \bigg)
\end{equation}
and by the Cauchy-Schwarz inequality, we see that the new variance is always smaller than $\sigma^2$, i.e. showing some sign of convergence. For example, when $\beta= 0.9$ the variance converges to $\sigma^2/21 $; when $\beta = 0.99$ the variance converges to $\sigma^2/199$.
Yet, unless $\beta = 1$, the distribution has non-zero variance at infinite $m$. As $\beta \to 1$ (and keeping $m \gg 1/\beta$), we obtain
\begin{equation}
    \lim_{\beta \to \infty} \lim_{m\to \infty} \mathcal{N}\bigg(\mu, \sigma^2  \frac{1 -\beta}{1 + \beta} \frac{1 + \beta^{m-1}}{1 - \beta^{m-1}} \bigg) =  \mathcal{N}\bigg(\mu, \frac{\sigma^2}{m-1} \bigg)
\end{equation}
which agrees with the result using simple averaging. We may also use this relation to define a relation to approximate the exponentially averaged RVs:
\begin{equation}
    m  = \frac{2}{ 1 - \beta},\quad\quad \beta = \frac{m-2}{m}
\end{equation}
When we cannot solve for the exponentially averaged RVs, we rely on this approximation to give qualitative understanding. Also notice the relation $\beta = \frac{m-2}{m}$, which is reminiscent of the optimal momentum rate $\frac{m-3}{m}$ derived by Nesterov \citep{nesterov1983method, da2018general}.

Now we also would like to compute our estimated variance in this way. Similarly, we define
\begin{equation}
    \Bar{U}_\beta :=  (1-\beta)\sum_{i=1}^m \beta^{m - i} X_i^2;
\end{equation}
and the distribution of $\Delta \theta$ is given by the distribution of the variable $T := \frac{\Bar{X}_{\beta1} -\mu}{\sqrt{U_{\beta_2}}}$. However, this distribution cannot be solved for analytically due to the effect of exponential averaging, and, as in \citep{liu2019variance}, we approximate the effect of exponential averaging by simple averaging.  This approximation is good when $\beta$ is close to $1$, which is indeed the case in practice. The default value of $\beta$ for Adam is $0.999$ \citep{journals/corr/KingmaB14_adam}, and this is the choice of the majority of works that uses Adam as optimizer, for RMSProp, the default value is $0.99$ \citep{Tieleman2012_rmsprop}. This means that the simple-average approximation should apply to most of the practical situations we are aware of\footnote{The smallest value of $\beta$ in use that the authors are aware of is $0.98$ in \citep{lan2019albert}}. 

\vspace{-1mm}
\vspace{-1mm}
\section{Dependent $X-U$ distributions}
\vspace{-1mm}\vspace{-1mm}
We have shown in proposition~\ref{prop: bound} that the distribution of $\Delta \theta$ cannot be understood unless the correlation is taken into account. The estimated mean $\Bar{X}$ and $U$ are correlated since they are often estimated using the same samples. This turns out to have important implication for the distribution of the variable $T= \frac{\Bar{X}}{\sqrt{U}}$. To start, we consider the distribution of a random variable $G$ (which approximates $U_\beta$),
\begin{equation}
    G = \frac{X_m^2}{\frac{1}{c}\sum_{i=1}^m X_i^2};
\end{equation}
where $c=m$ or $m-1$ depending on which normalization condition we use; notice that the expected value of $G$ gives the variance of $\Delta \theta$.Once we obtain the distribution of $G$, we may then take the square root and perform a transformation of RV to obtain the other related distributions. Note that G can be written as
\begin{equation}
    K := \frac{X_m^2}{X_m^2 + H}  := \frac{G}{c},
\end{equation}
where $H$, according to the discussion before, is a RV obeying the $\chi^2$ distribution with degree of freedom $m-1$, and $X_m^2$ is a $\chi^2$ distribution with degree of freedom $1$.  $H$ and $X_m^2$ are uncorrelated by definition. For convenience, we also define the special case $m=1$.

\begin{equation}
    G = \frac{X_1^2}{|X_1|} = sgn(X_1)
\end{equation}

We now derive the distribution for $G$. Write $X^2_m$ as $Z$, and we know that $Z$ and $H$ are independent, and the joint distribution is then 
\begin{equation}
    f(z, h) = f(z) f(h) = \frac{1}{\sqrt{2\pi z}} e^{-z/2} \times \frac{1}{2^\frac{m-1}{2}\Gamma( \frac{m-1}{2} )} h^{\frac{m-1}{2} -1} e^{-h/2} 
\end{equation}
we transform the variable from $(Z, H)$ to $(Z, K)$ and then integrate out $Z$ to obtain the distribution for $K$. The Jacobian for this transformation is
\begin{equation}
    \det \frac{\partial (z, h)}{\partial(z, k)} = \frac{z}{k^2}. 
\end{equation}
The joint distribution for $(z, k)$ is then (noticing $h = z(1-k)/k$)
\begin{align}
    f(z, k)  
    & = \frac{1}{2^{\frac{m}{2}} \sqrt{\pi} \Gamma(\frac{m-1}{2})}
    \frac{1}{k^2} \bigg( \frac{1-k}{k} \bigg)^{\frac{m-1}{2} - 1} \times z^{\frac{m-2}{2}} e^{-z/k}
\end{align}
we now integrate over $z$ to obtain
\begin{align}
    f(k) 
    &=  \frac{\Gamma(\frac{m}{2})}{\Gamma(\frac{m - 1}{2}) \Gamma(\frac{1}{2})} \times 
    \frac{k^{-\frac{1}{2}}}{1-k} (1-k)^{\frac{m-1}{2}}
\end{align}
or, equivalently,
\begin{equation}\label{eq: k-distribution}
    f_m(k) = \frac{1}{B(\frac{m-1}{2}, \frac{1}{2})} k^{-\frac{1}{2}}(1-k)^{\frac{m-1}{2}-1}
\end{equation}
This is the $Beta(\frac{1}{2}, \frac{m-1}{2})$ distribution. The expected value of $k$ is:
\begin{align}
    \mathbb{E}_{f_m} [K] 
    &= \frac{1}{m}
\end{align}
The mode for this distribution is $\frac{1}{m-2}$.

Now we transform $k \to g$ to get ($k= \frac{g}{c}$) to obtain the distribution for $g$. First, we let $c= m$
\begin{equation}\label{eq: g-distribution}
    f_m(g) =  \frac{\Gamma(\frac{m}{2})}{\Gamma(\frac{m - 1}{2}) \Gamma(\frac{1}{2})} \times \frac{(\frac{g}{m})^{-\frac{1}{2}}}{1 - \frac{g}{m}}\bigg( 1 - \frac{g}{m}\bigg)^{\frac{m-1}{2}} \times \frac{1}{m}
\end{equation}
and, by definition, $0\leq g\leq m$. From the result on the $Beta$ distribution, we have
\begin{equation}
    \mathbb{E}_{f_m}[G] = 1
\end{equation}
and the mode is simply $\frac{m}{m-2}$. This shows that the variance of $\Delta \theta$ is a constant in time. On the other hand, had we chosen $c=m-1$, then the expected value would then be $\frac{m-1}{m}$, while the mode be $\frac{m-1}{m-2}$. This agrees with experiment. See section~\ref{sec: exp} for detail.

This is the distribution we want to find. This can be analytically integrated over  to find its moments, which are the quantities we care about. For now, we focus on the interesting special cases for $m$. We first consider the limiting distribution as $m > \infty$, we obtain
\begin{align}
    f_\infty(g) = \lim_{m\to \infty } f_m(g) 
    &= \frac{1}{\sqrt{2\pi}} g^{-\frac{1}{2}} e^{-\frac{g}{2}}
\end{align}
which is simply a $\chi^2$ distribution with degree of freedom $1$, with expected value $1$ and variance $2$, which is expected. The more interesting limit is when $m$ is small. For $m=3$, we have
\begin{equation}
    f(g) = \frac{1}{2\sqrt{3}}g^{-\frac{1}{2}}
\end{equation}
For $m = 2$. We have 
\begin{equation}
    f_2(g) = \frac{1}{\pi} \frac{1}{\sqrt{\frac{g}{2}\left(1-\frac{g}{2}\right)}}
\end{equation}
which is a (shifted and rescaled) sinusoidal distribution. 
This means that, when $m=2$, $g$ is either very close to $0$ or very close to $1$. We might even go one step further to see what happens if $m\to 1^+$. For $0<g<1$, $\lim_{m\to 1^+} f_m(g)=0$, while for $g=1$ and $g=0$, the limit is not well-defined. This is a signature that the distribution is tending to a mixed delta-distribution proportional to $\delta(g-1) + \delta(g)$. 
The fact that this is not a well-defined limit suggests that there is a singularity in the variable $g$ when $m\to 1$, and some qualitative transition has happened as we approach the limit. For $m=1$, by definition we should obtain a delta distribution $\delta(x-1)$, which is what one expects. 

Now we are ready to obtain the distribution for $X_k = \sqrt{K}$. Applying the transformation rule to Equation~\ref{eq: k-distribution} (and extend the distribution from $[0,\sqrt{m}]$ to $[-\sqrt{m}, \sqrt{m}]$), we obtain
\begin{equation}
    f_m(x_k) = \frac{\Gamma(\frac{m}{2})}{\Gamma(\frac{m - 1}{2}) \Gamma(\frac{1}{2})}  (1-{x_k^2})^{\frac{m-1}{2} - 1};
\end{equation}
this is the predicted distribution of $\Delta \theta$. We plot the theoretical p.d.f. in Figure~\ref{fig: theo pdf}. We may perform another transformation of variable to see that this is still $Beta$ distribution. As expected in Proposition~\ref{prop: bound}, this distribution is bounded and sub-gaussian. Let $z_k = \frac{1 + x_k}{2}$, we see 
\begin{equation}
    f_m(z_k) = c  (1-{z_k})^{\frac{m-1}{2} - 1} z_k^{\frac{m-1}{2} - 1}
\end{equation}
where $c$ is the normalizing constant. This is a $Beta(\frac{m-1}{2}, \frac{m-1}{2})$ distribution, centered, and with a Gaussian asymptotic distribution.

Now we proceed to define $X_g = \sqrt{G}$ and extend the support from $\mathbb{R}^+$ to $\mathbb{R}$; we obtain
\begin{equation}\label{eq: main result}
    f_m(x_g) = \frac{1}{\sqrt{m}}\frac{\Gamma(\frac{m}{2})}{\Gamma(\frac{m-1}{2}) \Gamma(\frac{1}{2})} \frac{1}{1 - \frac{x_g^2}{m}}\bigg(1-\frac{x_g^2}{m}\bigg)^{\frac{m-1}{2}}
\end{equation}
Again, when $m\to \infty$, 
\begin{equation}
    f_\infty(x_g) = \frac{1}{\sqrt{2\pi}} e^{-\frac{x_g^2}{2}}
\end{equation}
which is the Gaussian distribution. The more interesting case is also the non-asymptotic case. When $m=2$, we have that
\begin{equation}
    f_2(x_g) = \frac{1}{\pi} \frac{1}{\sqrt{1 - x_g^2}}
\end{equation}
which is the sinusoidal distribution. When $m=3$,
\begin{equation}
    f_3(x_g) = \frac{1}{2\sqrt{2}}
\end{equation}
which is a uniform distribution supported on $[-\sqrt{2}, \sqrt{2}]$.

To summarize, we have derived the major theoretical result of this work.
\begin{theorem}
    Let $\{g_i\}_{i=1}^t$ be i.i.d. sampled from $\text{Normal}(0, 1)$, $\mu=0$, and $n_t= \sum_{i=1}^t g_i^2$, $c_n=\frac{1}{t}$, then the p.d.f. of $\Delta \theta$ is given by Equation~\ref{eq: main result}.
\end{theorem}

It is interesting to compare this with the $t$-distribution. While the $t$-distribution is unbounded, this distribution is bounded for any finite $m$. As $m\to \infty$, the two distributions converge to the same limiting Gaussian distribution with variance $1$. The variance for $T$ at finite $m$ is $\frac{m}{m-2}$, while that of the $|X_g|$ has variance $\frac{m-1}{m}$, and we obtain the relation
\begin{equation}
    Var[T] > 1 > Var[|X_g|]
\end{equation}
and the inequality becomes equality when $m\to \infty$. This shows that approximating $\frac{X_m}{\sqrt{\sum^m X_i}}$ by neglecting the correlation between the numerator and denominator overestimates the variance; it also predicts to have the wrong trend: while the actual distribution has decreasing variance as $m$ decreases, the $t$-distribution predicts increasing variance, with its variance diverging at $m=2$.

\begin{figure*}[t!]
    \centering
    \includegraphics[width=0.8\linewidth]{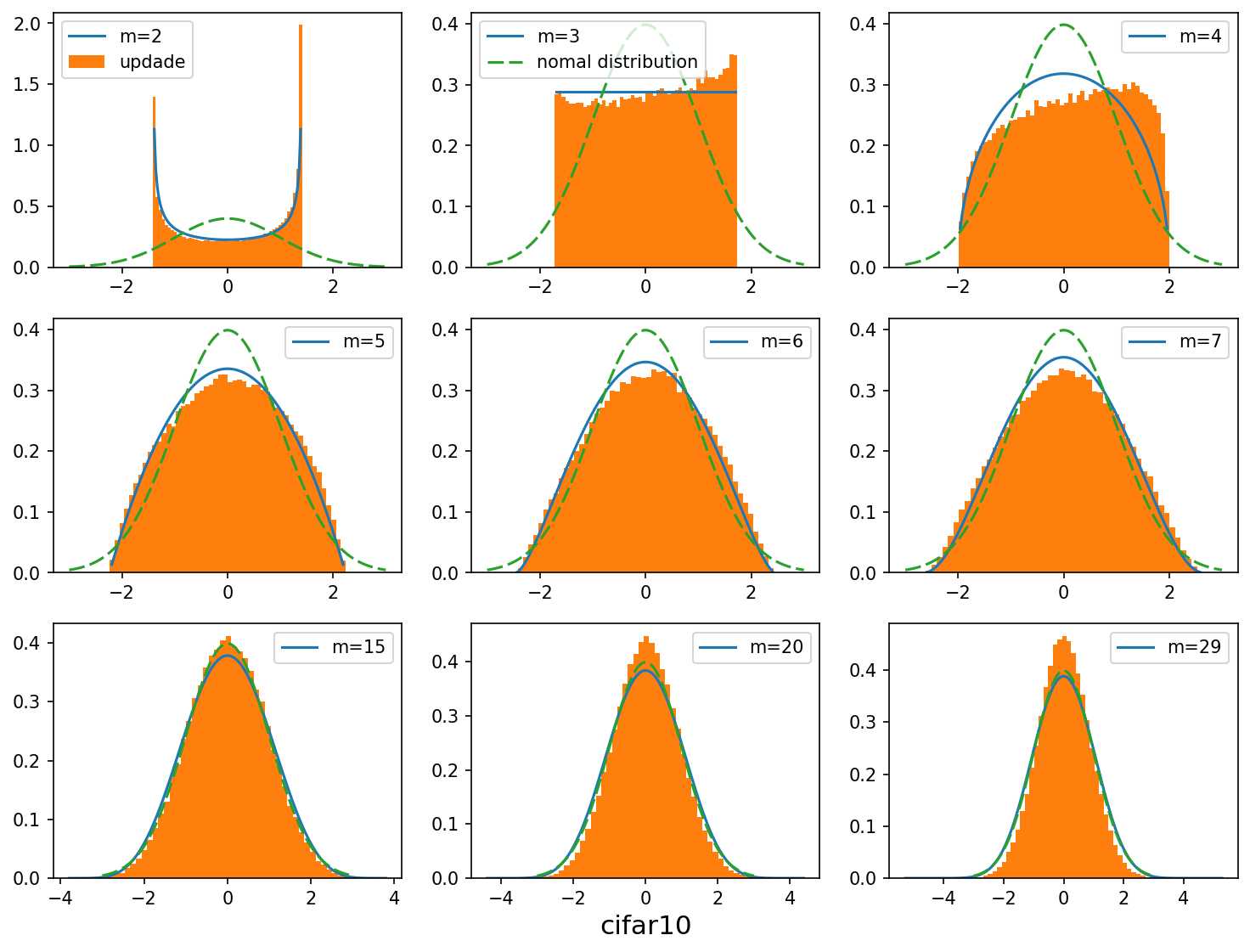}
    \vspace{-1em}
    \caption{Distribution of the update distribution of a non-hand-picked layer of a RegNetX-200MF trained on CIFAR-10. We see excellent agreement between our theory and experiments in the following sense: (1) the theoretical lines and the experimental histogram agree well visually and no outstanding disagreement exists; (2) the transition from a unimodal distribution to a bi-modal distribution occurs precisely at the same time step ($m=2$) as predicted ($m=2$). }
    \vspace{-1em}
    \label{fig:distribution for cifar10}
\end{figure*}

Besides the p.d.f. for $\Delta \theta$, the following two corollaries summarizes the most important predictions of this work. Agreement with the experiment would suggest that the assumptions made in this work are valid and applicable to real situations. The limitation is that these results hinge on the assumption that the gradient's underlying distribution is a zero-mean Gaussian with a constant variance; the deviation between the theoretical distribution and the measured distribution can be used to probe the skewed-ness of the gradient, the existence of a correlation between the different parameters, and the time evolution of the variance.
\begin{corollary}
The variance of $\Delta \theta$ is $1$.
\end{corollary}
\begin{corollary}
The variance of $|\Delta \theta|$ is 
\[1 - \frac{4 m \Gamma(m/2)^2}{(m-1)^2 \pi \Gamma(\frac{m-1}{2})^2}, \]
an increasing function of time and converges to a finite value as $m \to \infty$.
\end{corollary}

\vspace{-1mm}
\vspace{-1mm}
\section{Experiments}\label{sec: exp}
\vspace{-1mm}
In this section, we conduct experiments to test our theory. Notice that the experiments and plots are obtained from a single training trajectory instead of obtained by averaging over an ensemble of training trajectories with different initializations of the networks. We expect the agreement for all experiments to get even better when such ensembling is used. The fact that the agreement is right on a single trajectory level makes our theory more applicable to real problems (so that the practitioners only have to run once and check what went wrong). The agreement is expected to become better if we average over multiple runs.

\begin{figure*}
    \centering
    \includegraphics[width=1.0\linewidth]{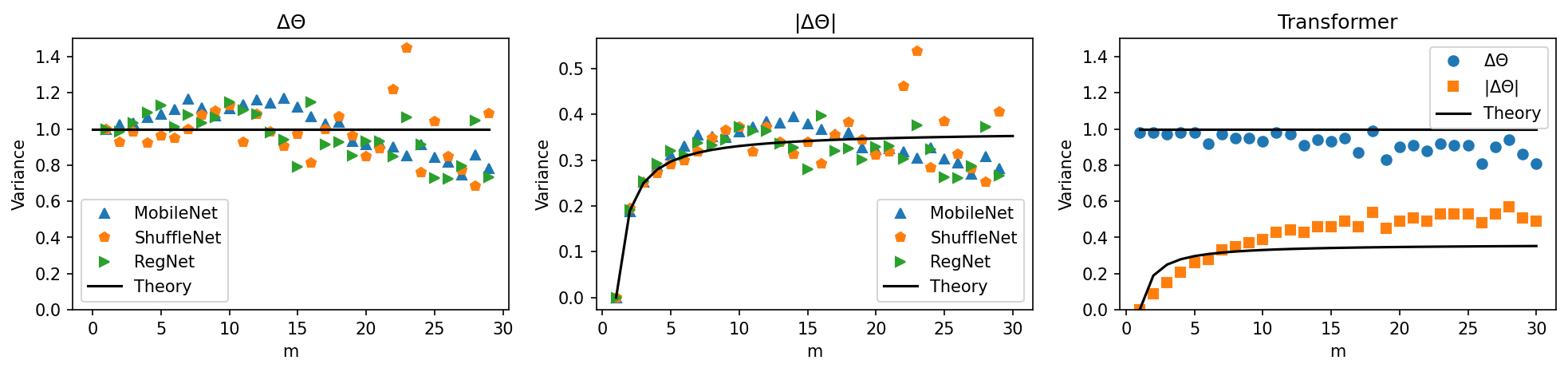}
    \vspace{-1em}
    \caption{\textbf{Left}: variance of $\Delta \theta$. \textbf{Middle}: variance of $|\Delta \theta|$. \textbf{Right}: variance of the transformer. We see that for both datasets, the agreement between our theory and experiment is excellent. There are two points worth noticing: (1) for both dataset, the agreement is very good for $m\leq 10$, suggesting that a constant Gaussian distribution well approximates the distribution of the gradient before this point; (2) the agreement of RegNetX on CIFAR-10 continues to hold much later, while the smaller network on MNIST starts to deviate after $m>10$. (3) The transformer also agrees well with the predictiong, showing well-behaving and bounded variance in the update.}
    \label{fig: variance theory and experiment}
\end{figure*}

\begin{figure*}[h]
    \centering
    \includegraphics[width=0.78\linewidth]{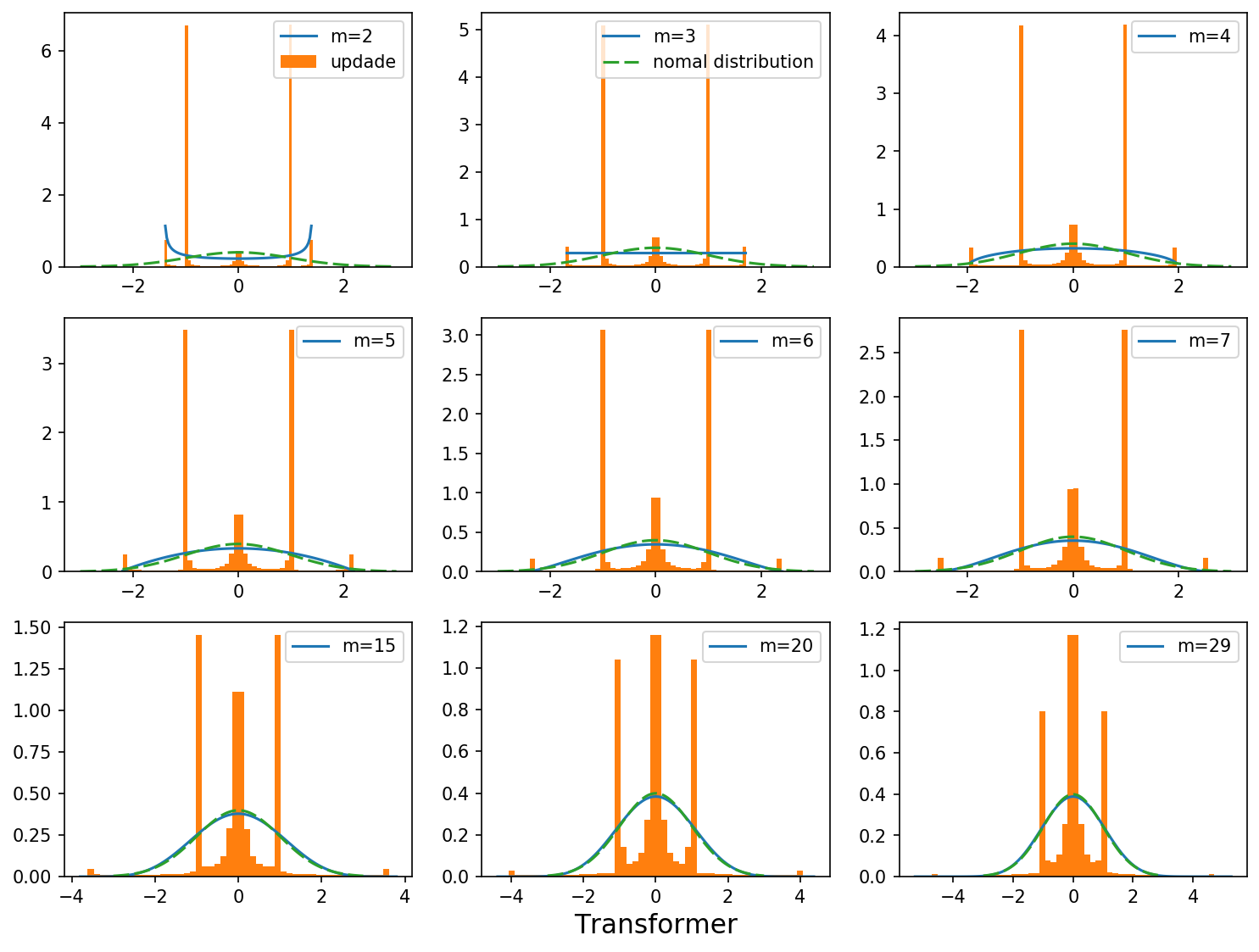}
    \vspace{-1em}
    \caption{Distribution of the update distribution of transformer on IWSLT14 Ger-Eng. We see that the distribution of the updates of a transformer is quite pathological, defying any simple characterization; for example, there seems to be at least $5$ prominent peaks from step $1$ to $20$.}
    \vspace{-1em}
    \label{fig:distribution for transformer}
\end{figure*}
\vspace{-1mm}
\vspace{-1mm}
\subsection{Distribution the Update}
\vspace{-1mm}
We measure the probability density function of the update of the adaptive gradient when we train on RegNetX-200MF \citep{radosavovic2020designing} on the CIFAR-10 dataset\footnote{RegNetX is the newest generation of modern CNN architectures with state-of-the-art performance in computer vision. It is based on the residual structure \citep{he2016deep}}. See Figure~\ref{fig:distribution for cifar10}. We see that the agreement between our prediction and experiment is good both quantitatively and qualitatively. Qualitatively, the following two predictions are confirmed: (1) the distribution transitions from a bi-modal distribution to a uni-modal distribution at $m$ increases; for this task, it is even more surprising that the transition to a uni-modal distribution occurs exactly at the predicted time step $m=3$; (2) the distribution converges to a gaussian distribution. A similar level of agreement is observed for all other layers of the network (see appendix); this suggests our assumptions' general applicability and, therefore, our theory. One interesting point is that the empirical data seems slightly right-skewed. We hypothesize that this is because the underlying distribution of $g_t$ has a non-vanishing mean (also see our experiments in Section~\ref{sec: what affects distribution} for why this is the case). In the appendix, we also plot the single-trajectory update distribution for VGG \citep{simonyan2014very}, ResNet-18 \citep{he2016deep}, ShuffleNet-V2 \citep{zhang2018shufflenet}, ResNeXt-29 \citep{xie2017aggregated}, MobileNet \citep{howard2017mobilenets}, EfficientNet-B0 \citep{tan2019efficientnet}, and DenseNet-121 \citep{huang2017densely}, which we show to also agree well with the prediction\footnote{Some deviation is also observed, but such deviations seem to appear in a consistent way for a fixed architecture, suggesting that the actual distribution has some sensitivity to the model.}. 

We also plot compare the distribution over different random initializations in Figure~\ref{fig:different seeds}. We plot the overlap of three different random seed in dark red, which also agrees well with the prediction. We see that the variance of the distribution across different seeds is relatively small, and all agree well with the theoretical prediction.


\begin{figure}
    \centering
    \includegraphics[width=0.4\linewidth]{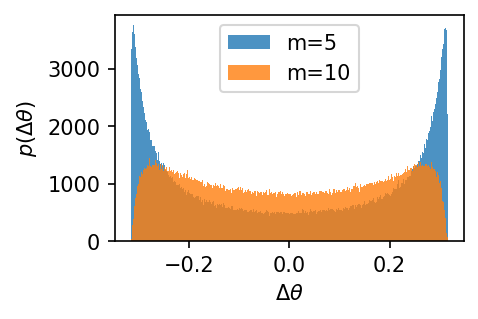}
    \includegraphics[width=0.4\linewidth]{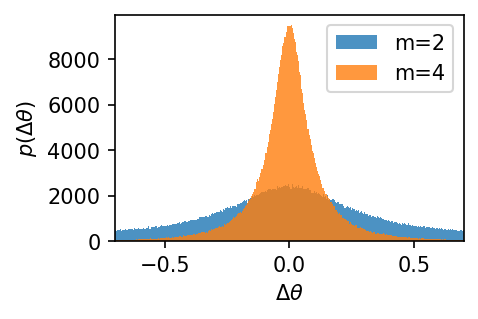}
    \vspace{-1em}
    \caption{\textbf{Left}: $g_t \sim \text{Normal}(0, t^2)$. \textbf{Right}: $g_t \sim \text{Normal}(0, \frac{1}{t^2})$. We see that, when the expected magnitude of the gradient is decreasing, the distribution of the update quickly becomes uni-model; when the magnitude of $g_t$ is increasing, the bi-modal structure is retained for much longer. }
    \label{fig:time dependent variance}
\end{figure}
\vspace{-1mm}
\vspace{-1mm}
\subsection{Increasing and Bounded variance}
\vspace{-1mm}
One of the key predictions of this work is that the adaptive gradients' variance is an increasing function of time when the training is not far from initialization, and we verify this prediction in this section. We plot the variance of the update versus the time step $m$ on CIFAR-10, for different model architectures: RegNetX, MobileNet \citep{howard2017mobilenets}, and ShuffleNet \citep{zhang2018shufflenet}. See Figure~\ref{fig: variance theory and experiment}. The agreement between our theory and experiment is excellent. There are two points worth noticing: (1) the agreement is very good for $m\leq 20$, suggesting that a constant Gaussian distribution well approximates the distribution of the gradient before this point; (2) there is no occasion when the variance of the update becomes anomalously large. This agreement suggests that (1) the distribution of gradient at initialization can be well-approximated by a time-independent gaussian at the initialization; (2) how far this approximation continues to hold as $m$ increases depends on the architecture of the model and the nature of the task. In the appendix, we show that there are also architectures that deviate from the predicted value from smaller values of $m$, but they all have smaller variances than the predicted value, which further corroborates with the intended message that there is no problem with the variance of the adaptive gradients.

\subsection{Distribution of the Transformer}
In this section, we study the distribution of the original transformer architecture \citep{vaswani2017attention}, since this is the situation where the distribution of the adaptive gradients is pathological. As in \citep{liu2019variance}, we run transformer on the IWLST14 Ger-Eng dataset, using the default implementation in the Fairseq package \citep{ott2019fairseq}. See Figure~\ref{fig: variance theory and experiment}-Right. We see that the transformer also has a very regular variance in the update, as expected by the theory. This suggests that the attribution of the pathological training behavior of transformers to the variance of Adam \citep{liu2019variance} is incorrect. We now show the empirical distribution of the updates in a single layer of the transformer architecture in Figure~\ref{fig:distribution for transformer}. We see that the distribution of the transformer is pathological and does not seem to obey a simple distribution. For example, at step $5$ ($m=5$), there seem to be five prominent peaks in a single layer, which is a sign that the distribution is a composite one. Comparing with the experiment in Figure~\ref{fig:time dependent variance}.LEFT and Figure~\ref{fig:cauchy}, we hypothesize that there are at least two different underlying distributions: (1) a heavy-tail-like distribution feature the first, third, and fifth peak, and (2) a distribution with increasing variance featuring the second and the fourth peak. One essential future work is to investigate the cause of each of these peaks, which we believe, will shed new and important light on this poorly understood problem.

\vspace{-1mm}
\vspace{-1mm}
\section{What affects the distribution of the update?}\label{sec: what affects distribution}
\vspace{-1mm}
During training, it is often the case that the practitioners have to continually check the distribution of the gradient to infer what might be wrong with the training. For example, when the training of a model does not proceed well; one might suspect that it is due to (1) update too small; (2) update too large; or (3) update fluctuates too much, and these can only be known once one inspects the distribution of the update.
Therefore, it is worth studying what might affect the distribution of the update. In our framework, this effect can incorporated by setting the variance of the underlying gradient distribution to be a function of time: $g_t \sim \text{Normal}(0, \sigma^2(t))$. The distribution of the corresponding update $\Delta \theta$ no longer takes a simple analytic form now; therefore, we resort to simulations to answer this problem.

\vspace{-3mm}
\subsection{When is the update distribution bi-modal?}
\vspace{-1mm}
It is identified in \citep{liu2019variance} that the number of modes in the update distribution might affect ease of optimization, and the authors incorrectly attributed the cause to the divergence of the variance of the update. In this section, we show that the cause of bi-modality is the sudden increase of the variance of the gradient. See Figure~\ref{fig:time dependent variance}. We experiment with the following two kinds of time-dependent gradient distribution: (1) $g_t \sim \text{Normal}(0, t^2)$; (2) $g_t \sim \text{Normal}(0, \frac{1}{t^2})$.
We see that, in an increasing variance, the bi-modal structure remains for a relatively long time, while, in a decreasing variance, the bi-modal structure disappears very fast, and the distribution becomes thinner and sharper through time. This deformity can be compared with some real-task update distributions we encounter. For example, we plot the distribution of training a simple $4$-layer CNN on MNIST in the appendix, where the bi-modal structure is shown to remain for a relatively long time; this suggests that, in this task, the variance of the gradient is increasing as time proceeds.

\begin{figure}
    \centering
    \includegraphics[width=0.48\linewidth]{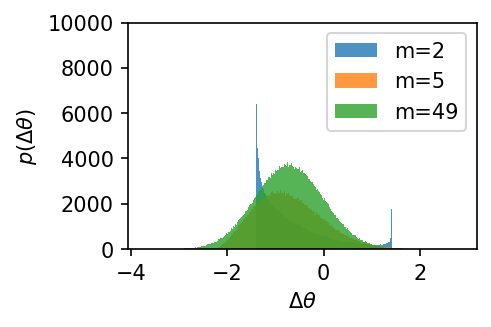}
    \includegraphics[width=0.48\linewidth]{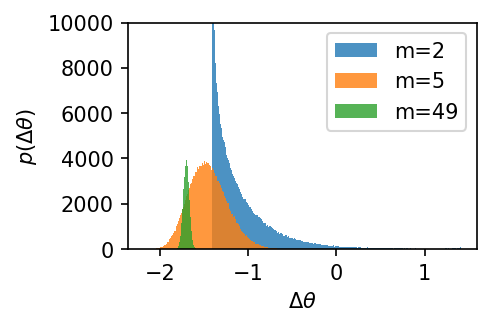}
    \vspace{-1em}
    \caption{\textbf{Left}: $g_t \sim \text{Normal}(-1, 1)$. \textbf{Right}: $g_t \sim \text{Normal}(-t, 1)$. We see that if $g_t$ obeys a non-zero mean distribution, the resulting distribution of $\Delta \theta$ is skewed, with initial steps more skewed than the asymptotic steps.}
    \label{fig: skewed}
\end{figure}

\vspace{-1mm}
\vspace{-1mm}
\subsection{When is the distribution skewed?}
\vspace{-1mm}
Intuitively, the answer is simple: when the underlying distribution of $g_t$ has non zero-mean. Here, we do some numerical simulation to check what that will affect the distribution quantitatively. Here, we experiment with following two kinds of noise; since the distribution simply flips if we change the sign of the mean, we only simulate with negative mean value: (1) $g_t \sim \text{Normal}(-1, 1)$; (2) $g_t \sim \text{Normal}(-t, 1)$.
See Figure~\ref{fig: skewed} for the result. We see that the resulting distribution is left-skewed but moves back towards the center as $m$ increases. It is also interesting to notice that the distribution gets thinner and less skewed when the magnitude of the mean increases with time.

\begin{figure}
    \centering
    \includegraphics[width=0.9\linewidth]{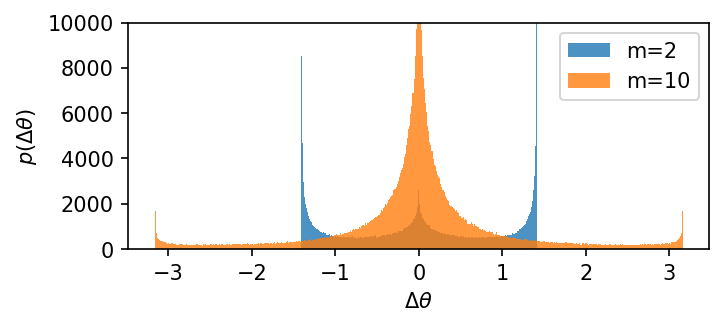}
    \vspace{-1.5em}
    \caption{$g_t$ is drawn from a standard Cauchy distribution. Interestingly, the distribution takes a trimodal structure.}
    \label{fig:cauchy}
\end{figure}

\vspace{-1mm}
\vspace{-1mm}
\subsection{Heavy-tailed gradient distribution}
\vspace{-1mm}
The gradient may take a heavy-tailed distribution\footnote{There is increasing research interest in studying the case when the underlying gradient is heavy-tailed \citep{gurbuzbalaban2020heavy, simsekli2019tail}. It is not rare that the update distribution of real tasks shows a trimodal structure. Many figures in \citep{liu2019variance} suggest the existence of a trimodal structure when a transformer is trained on a machine learning task.}. We simulate this by sampling $g_t$ from a standard Cauchy distribution. See Figure~\ref{fig:cauchy}. Interestingly, the distribution takes a trimodal structure. This trimodality suggests that the update distribution's trimodal structure may suggest (1) the existence of a large-norm but sparse gradient or (2) the training has very strong or even divergent noise. In fact, the update distribution for ResNet-18 and VGG21 are observed to be clearly bimodal at the initial step while the other studied models do not; see the appendix.

\vspace{-1mm}
\vspace{-1mm}
\section{Concluding Remark}
\vspace{-1mm}
In this work, we studied the distribution of the update of the adaptive gradient methods. We showed that the variance of (any higher moment of, in fact) of the adaptive gradient method does not diverge. The variance of the update is a constant, and the variance of its magnitude is an increasing function of time; this, in turn, means that the adaptive gradient methods are surprisingly stable, especially at the beginning of training. Adam's unexpected stability at initialization also implies that the previous understanding that the need for the warmup of Adam \citep{liu2019variance} is still an open problem. We believe that identifying the correct cause of Adam's incapability to train transformers will ultimately benefit the community.

\textit{Implications.} There are a few important implications of this work. \textbf{Correlation of the gradient.} Our analysis's limitation is obvious, the most important one being that the gradient is assumed to be uncorrelated for different parameters; this assumption leads to the analytical formula we derived, which, intuitively, is unlikely to be the case for a non-linear system. What is surprising here is that a theory based on such limited theory agrees excellently with the experiments; this, in turn, implies that the gradient noises of minibatch SGD are likely well-approximated by a diagonal matrix. The reason behind this anomalous suppression of off-diagonal covariance should be a crucial topic to study in the future. \textbf{Approximate Bayesian Inference.} Knowing the posterior distribution of the iteration of the optimization algorithm has important application to approximate Bayesian inference, where SGD and Adam are used as MCMC algorithms for approximating the posterior distribution of the model parameters \citep{mandt2017stochastic, liu2020stochastic, ziyin2021minibatch}, which are often useful for assessing the uncertainty in the model parameters.





\appendix
\onecolumn

\section{Additional Experiments}


\subsection{Distribution across different random seeds}
\begin{figure}[h]
    \centering
    \includegraphics[width=0.6\linewidth]{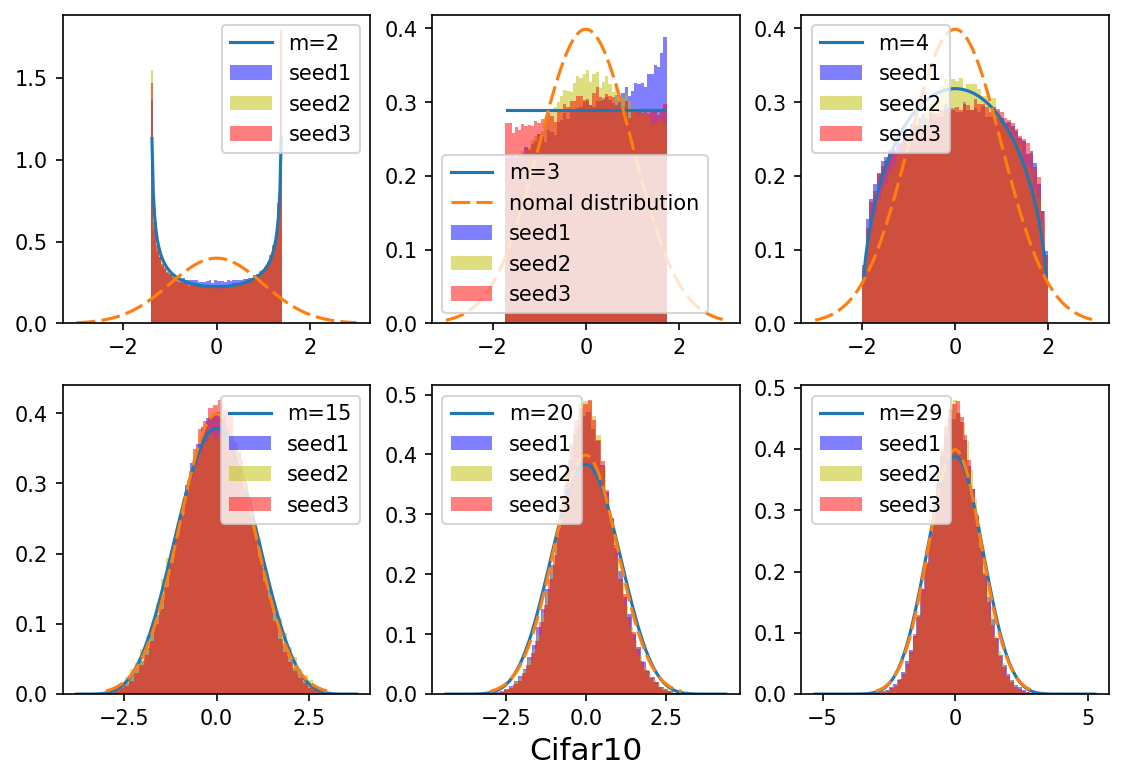}
    \vspace{-2em}
    \caption{Update distribution for different random seeds.}
    \label{fig:different seeds}
\end{figure}

We also plot compare the distribution over different random initializations in Figure~\ref{fig:different seeds}. We plot the overlap of three different random seed in dark red, which also agrees well with the prediction. We see that the variance of the distribution across different seeds is relatively small, and all agree well with the theoretical prediction. 

\subsection{Other Experiments on CIFAR-10}
\begin{figure}[b!]
    \centering
    \includegraphics[width=\linewidth]{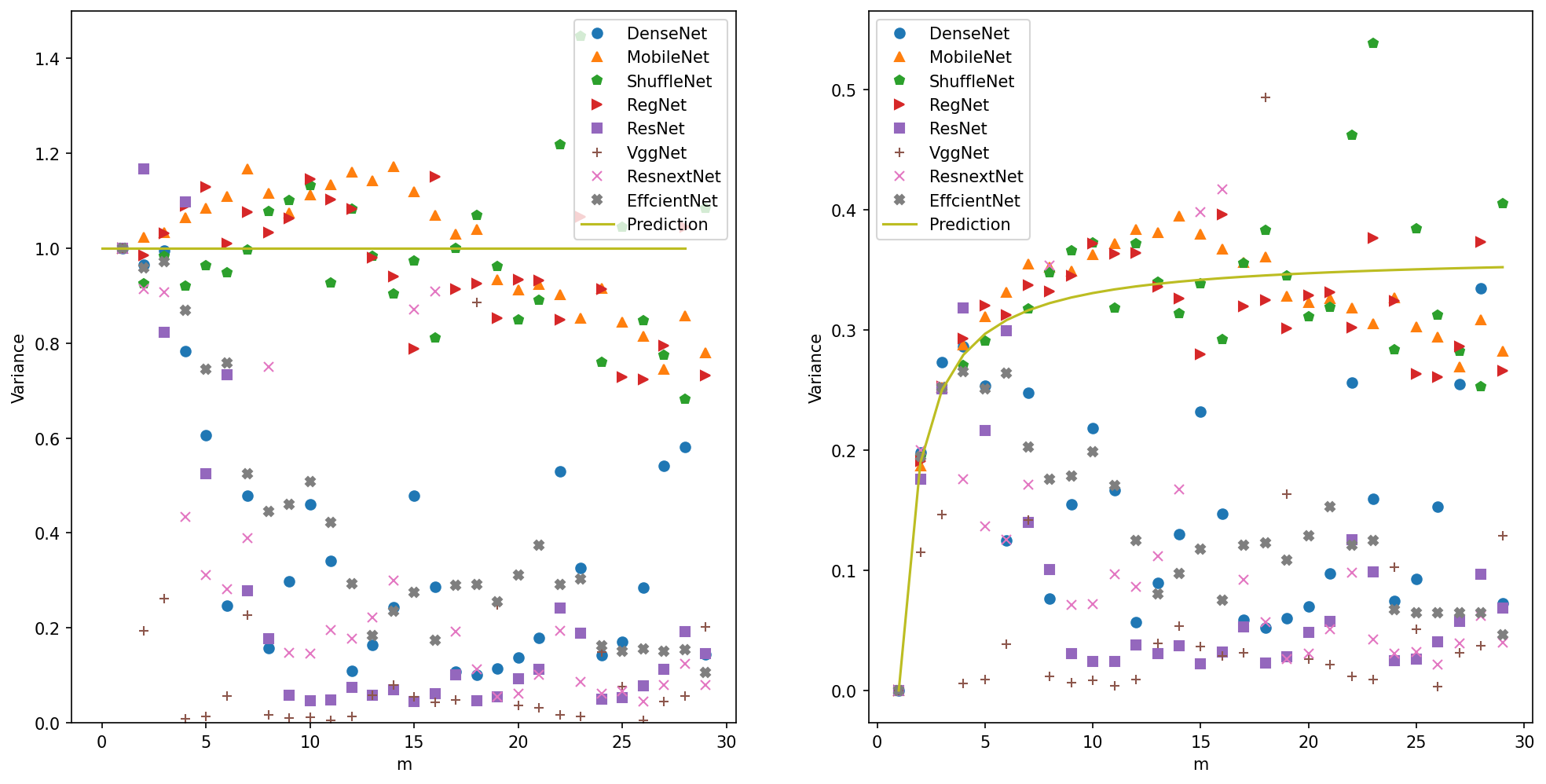}
    \caption{Left: variance of $\theta$. Right: variance of $|\theta|$. We see that, some architectures agree quite well our analysis, while other architectures deviates in a rather consistent way.}
    \label{fig:variances }
\end{figure}

In this section, we present more experiments to validify our theory. We first plot the distribution of update for a different layer (from what appeared in the main text) RegNetX-200MF trained on CIFAR-10, and excellent agreement between our theory and experiment is observed. We then plot the distribution of updates from random layers (with more than $10^4$ parameters) of VGG \citep{simonyan2014very}, ResNet-18 \citep{he2016deep}, ShuffleNet-V2 \citep{zhang2018shufflenet}, ResNeXt-29 \citep{xie2017aggregated}, MobileNet \citep{howard2017mobilenets}, EfficientNet-B0 \citep{tan2019efficientnet}, DenseNet-121 \citep{huang2017densely}. We first plot the variance of $\theta$ and $|\theta|$ in Figure~\ref{fig:variances }. While some nets deviate from the theory, others agree quite well. For those that deviate from the prediction, we observe that they all have smaller variances than the predicted value, this agrees with our message that the adaptive gradient methods do not have exploding variance problem at the beginning of training. We then plot the distributions of $\theta$ for a single trajectory for each of these models.

\begin{figure*}[h]
    \centering
    \includegraphics[width=0.95\linewidth]{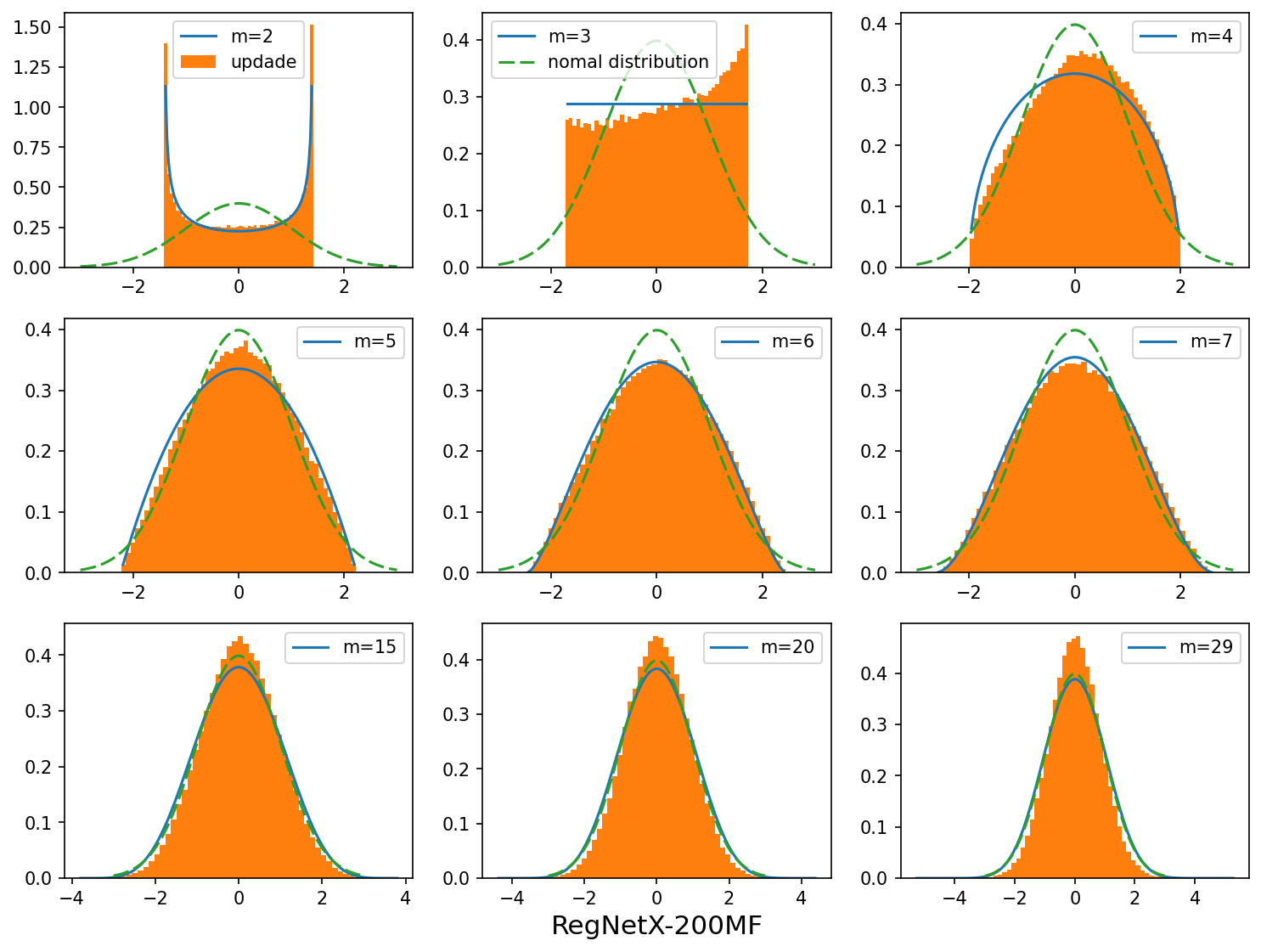}
    \caption{Distribution of the update distribution of another non-hand-picked layer of a RegNetX-200MF trained on CIFAR-10.  }
\end{figure*}

\begin{figure*}[h]
    \centering
    \includegraphics[width=0.75\linewidth]{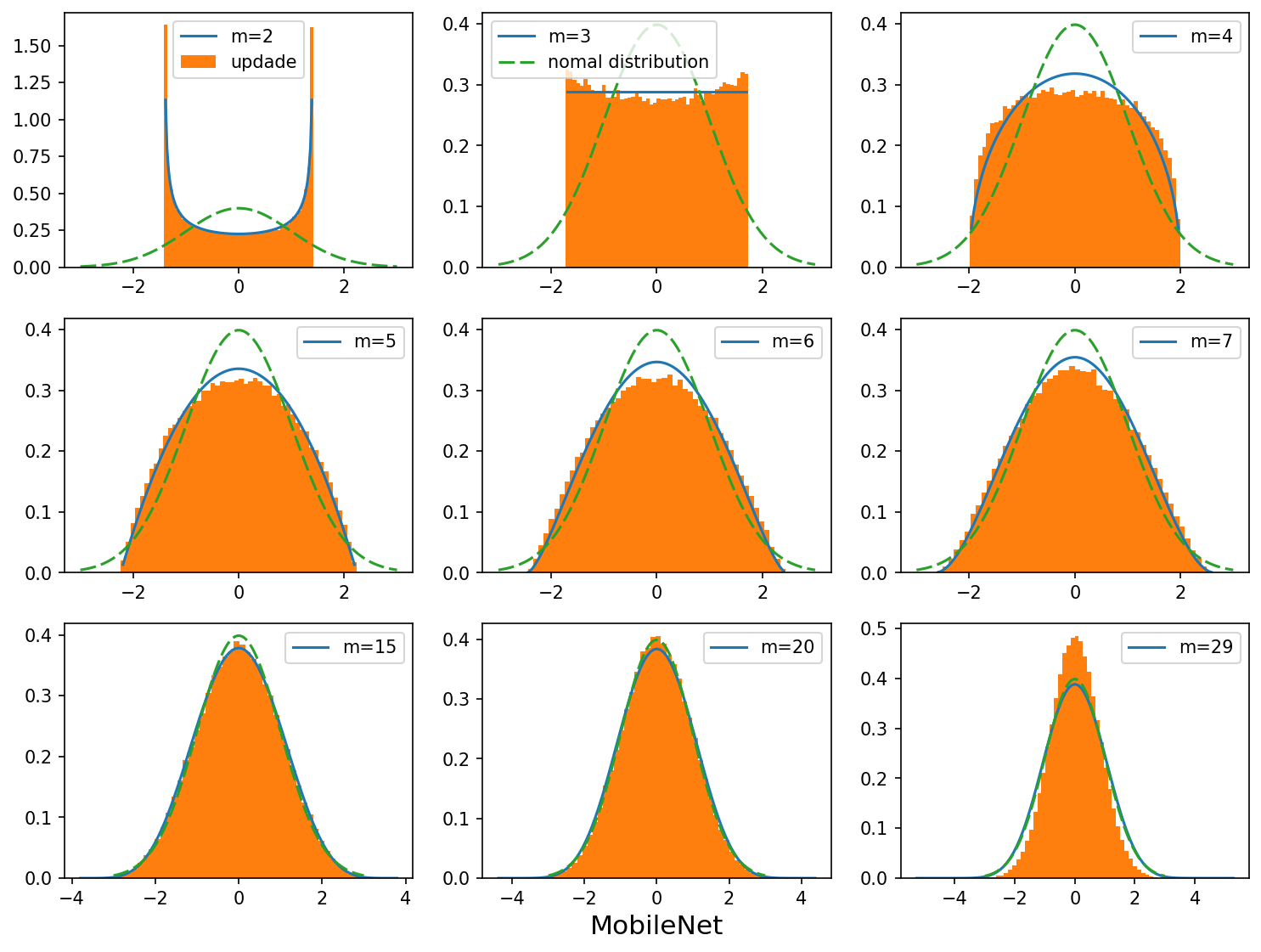}
    \caption{Distribution of the update distribution of another non-hand-picked layer of a MobileNet trained on CIFAR-10.  }
\end{figure*}

\begin{figure*}[h]
    \centering
    \includegraphics[width=0.75\linewidth]{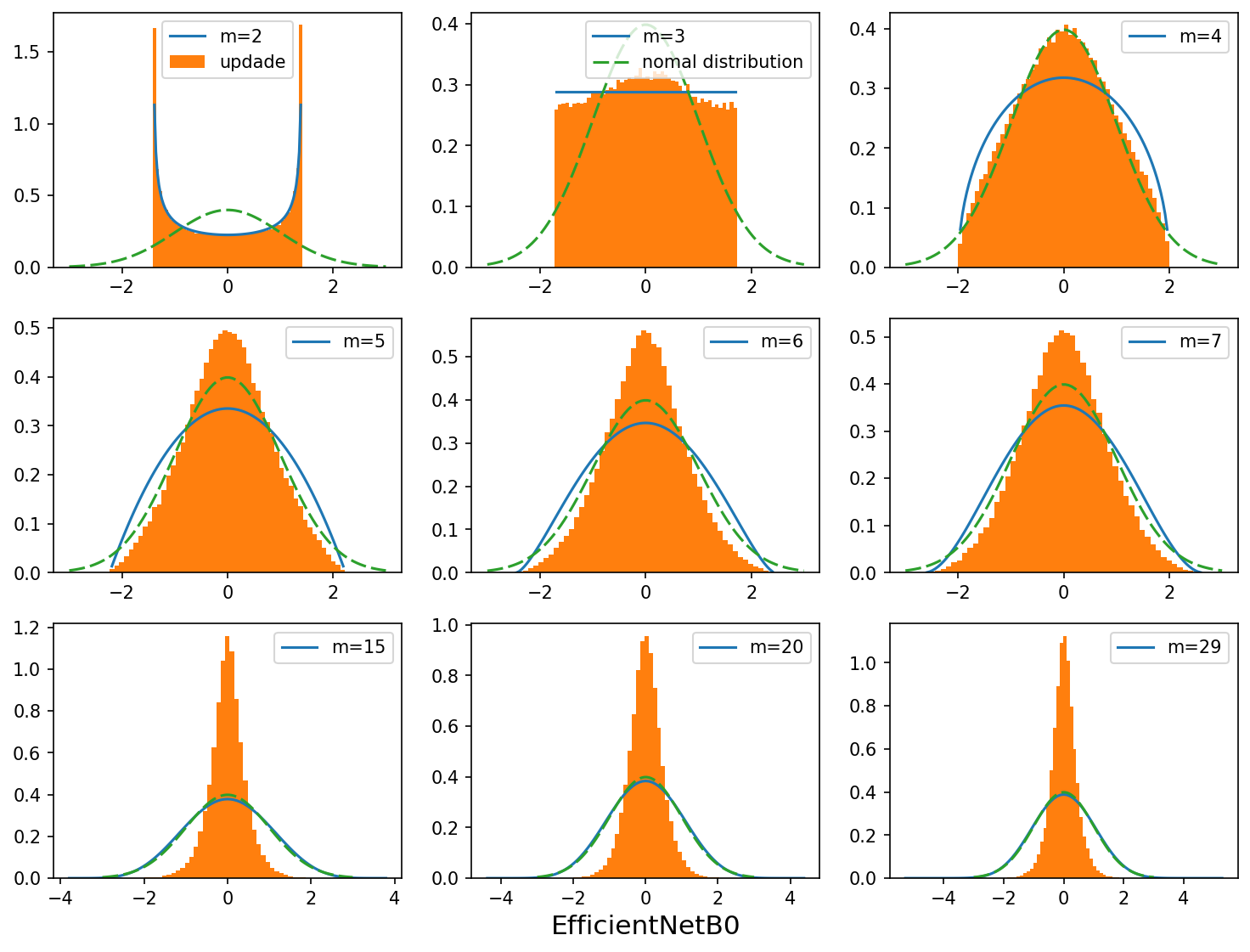}
    \caption{Distribution of the update distribution of another non-hand-picked layer of a EfficientNet trained on CIFAR-10.  }
\end{figure*}

\begin{figure*}[h]
    \centering
    \includegraphics[width=0.75\linewidth]{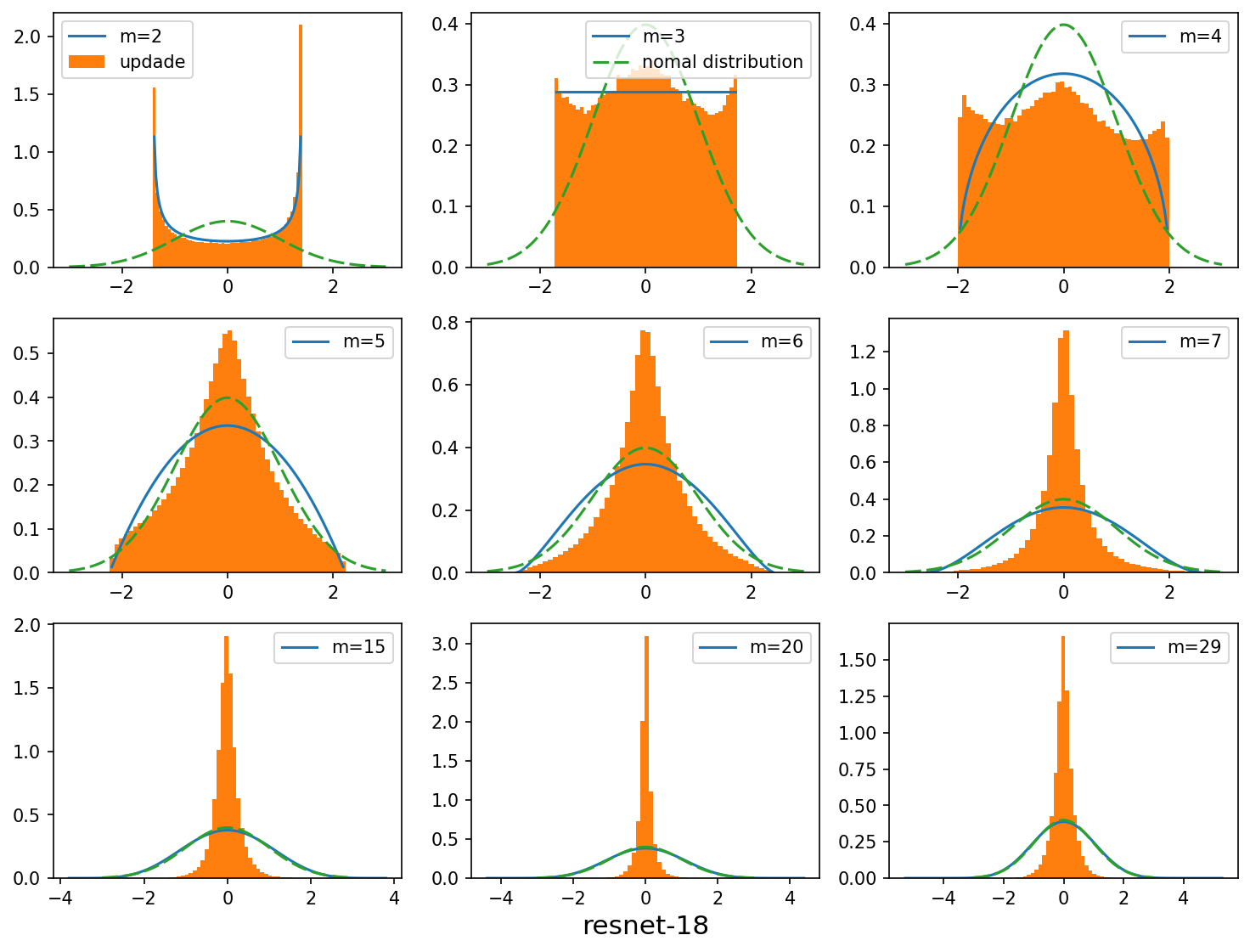}
    \caption{Distribution of the update distribution of another non-hand-picked layer of a ResNet-18 trained on CIFAR-10.  }
\end{figure*}

\begin{figure*}[h]
    \centering
    \includegraphics[width=0.75\linewidth]{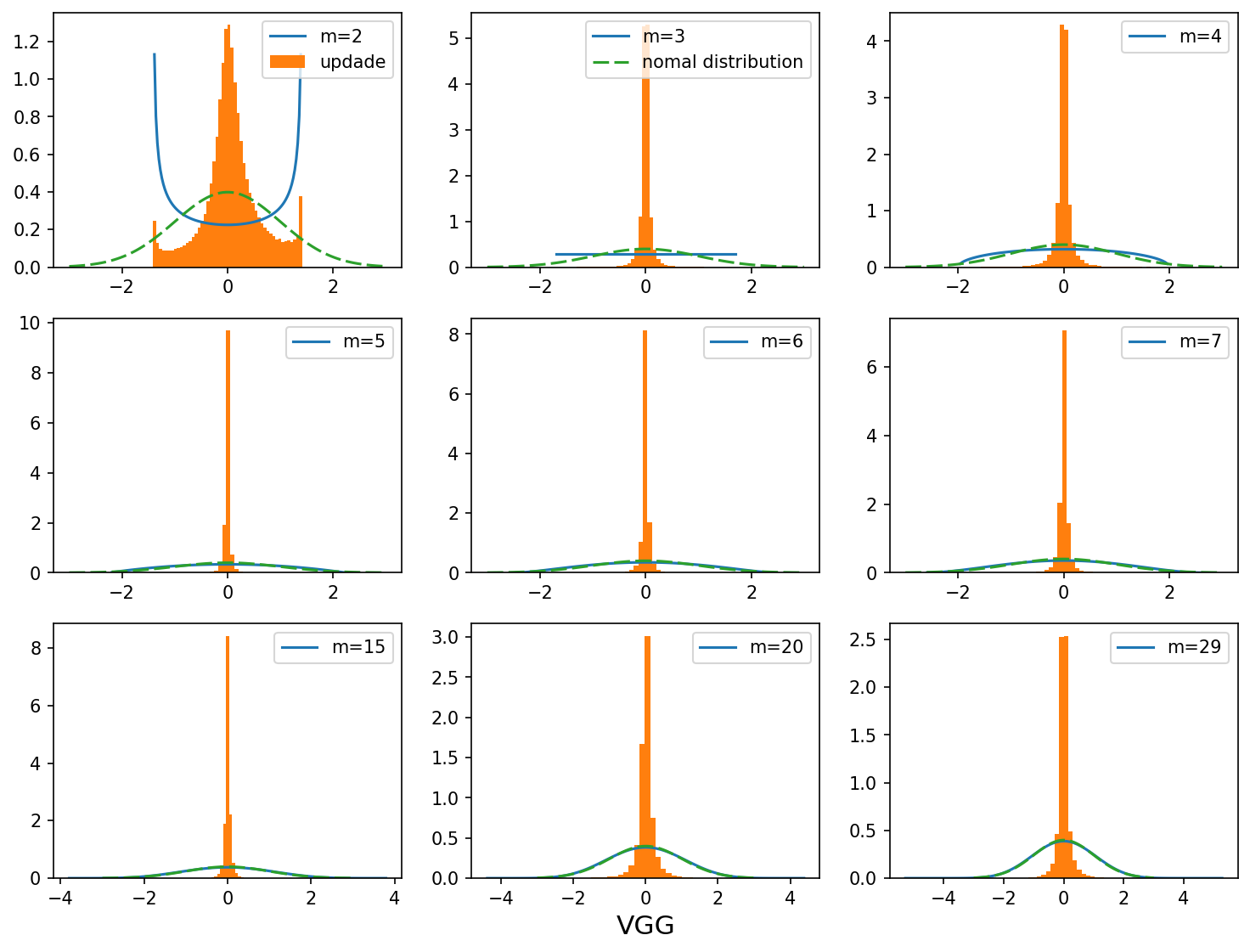}
    \caption{Distribution of the update distribution of another non-hand-picked layer of a VGG trained on CIFAR-10.  }
\end{figure*}

\begin{figure*}[h]
    \centering
    \includegraphics[width=0.75\linewidth]{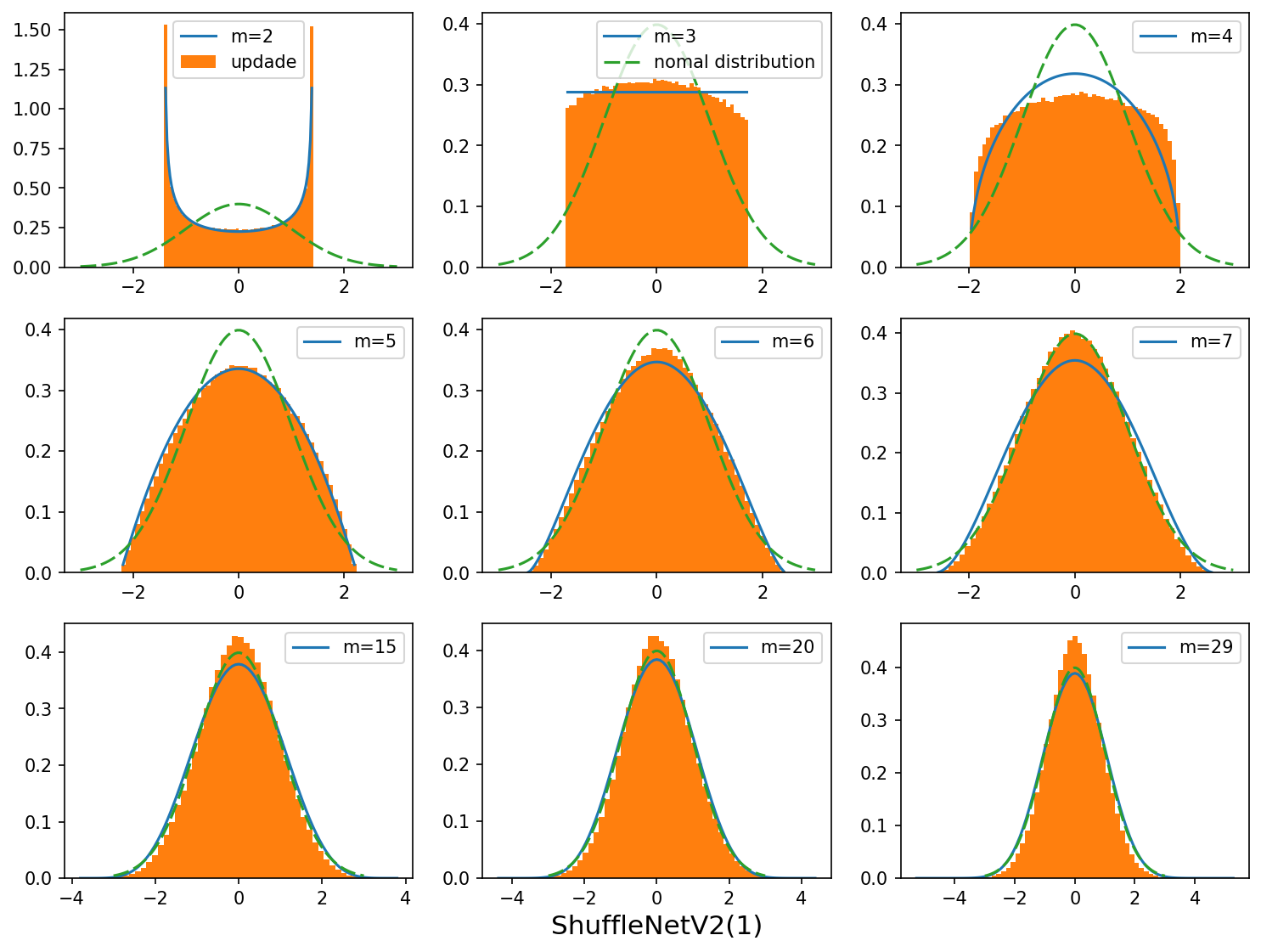}
    \caption{Distribution of the update distribution of another non-hand-picked layer of a ShuffleNet trained on CIFAR-10.  }
\end{figure*}

\begin{figure*}[h]
    \centering
    \includegraphics[width=0.75\linewidth]{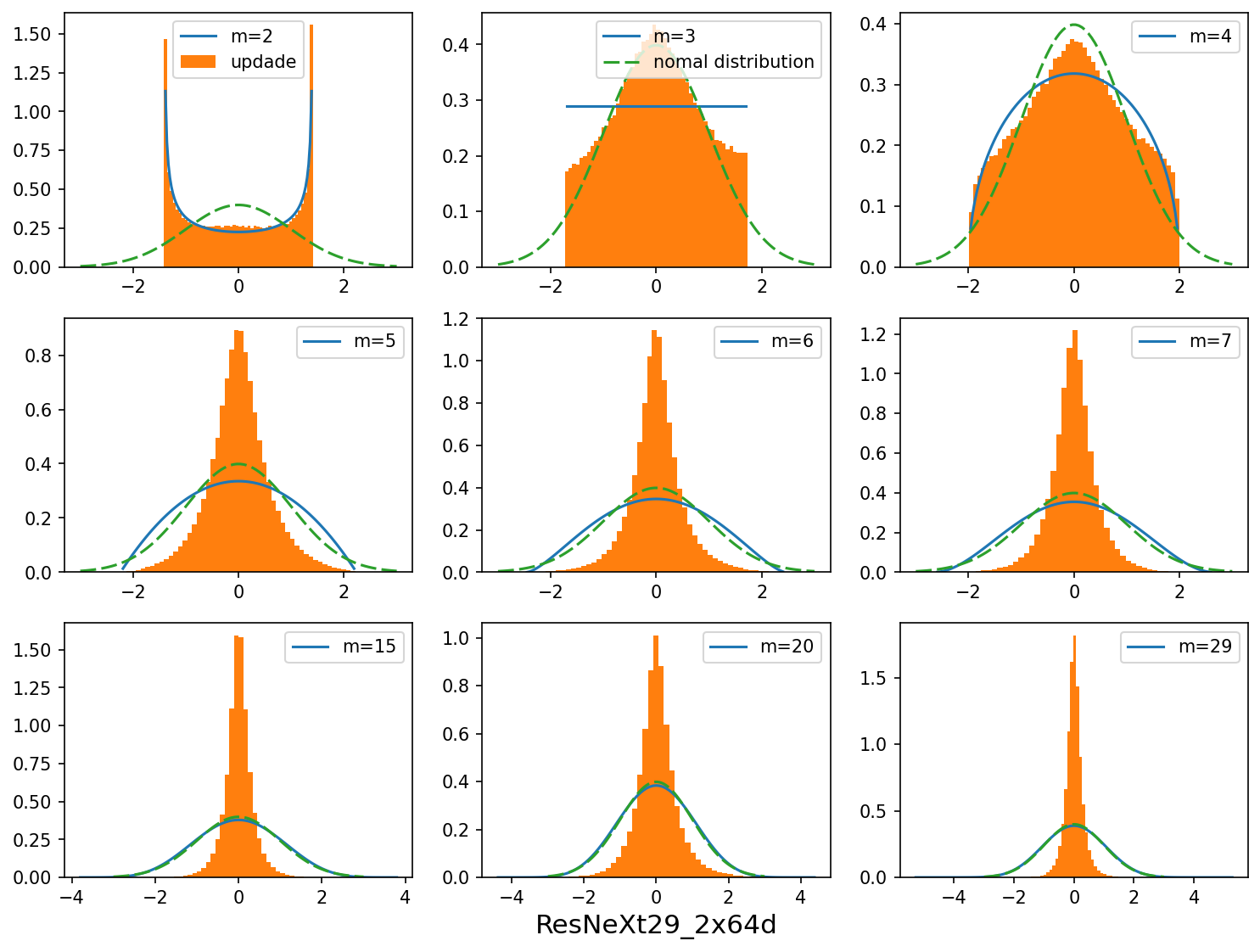}
    \caption{Distribution of the update distribution of another non-hand-picked layer of a ResNeXt29 trained on CIFAR-10.  }
\end{figure*}



\end{document}